\documentclass{article}

\usepackage[preprint,nonatbib]{neurips_2025}

\usepackage[utf8]{inputenc} 
\usepackage[T1]{fontenc}    
\usepackage[colorlinks,citecolor=green]{hyperref}       
\usepackage{url}            
\usepackage{booktabs}       
\usepackage{amsfonts}       
\usepackage{nicefrac}       
\usepackage{microtype}      
\usepackage{xcolor}         
\usepackage[numbers]{natbib}
\usepackage{pifont}
\usepackage{bbding}
\usepackage{}
\usepackage{algorithm}
\usepackage{algorithmic}
\usepackage{multirow}
\usepackage[inline]{enumitem}
\usepackage{enumitem}
\usepackage{boldline}
\usepackage{hhline}
\usepackage{amsmath}
\usepackage{graphicx}
\usepackage{pifont}

\usepackage{wrapfig}

\newcommand\ie{\emph{i.e.}}

\newcommand\eg{\emph{e.g.}}

\title{Task-Core Memory Management and Consolidation for Long-term Continual Learning}

\author {
    Tianyu Huai\textsuperscript{\rm 1},
    Jie Zhou\textsuperscript{\rm 1}\thanks{Corresponding author, jzhou@cs.ecnu.edu.cn.},
    Yuxuan Cai\textsuperscript{\rm 2}, 
    Qin Chen\textsuperscript{\rm 1},
    Wen Wu\textsuperscript{\rm 1},
    Xingjiao Wu\textsuperscript{\rm 1}, \\
    \textbf{Xipeng Qiu}\textsuperscript{\rm 3}, 
    \textbf{Liang He}\textsuperscript{\rm 1} \\
    \textsuperscript{\rm 1}School of Computer Science and Technology, East China Normal University \\
    \textsuperscript{\rm 2}School of Electrical and Electronic Engineering, Nanyang Technological University \\ 
    \textsuperscript{\rm 3}Computation and Artificial Intelligence Innovative College, Fudan University \\
}

\begin{document}

\maketitle

\begin{abstract}
In this paper, we focus on a \textbf{long-term continual learning} (CL) task, where a model learns sequentially from a stream of vast tasks over time, acquiring new knowledge while retaining previously learned information in a manner akin to human learning. Unlike traditional CL settings, long-term CL involves handling a significantly larger number of tasks, which exacerbates the issue of catastrophic forgetting. Our work seeks to address two critical questions: 1) How do existing CL methods perform in the context of long-term CL? and 2) How can we mitigate the catastrophic forgetting that arises from prolonged sequential updates?
To tackle these challenges, we propose a novel framework inspired by human memory mechanisms for long-term continual learning (\texttt{Long-CL}). Specifically, we introduce a task-core memory management strategy to efficiently index crucial memories and adaptively update them as learning progresses. Additionally, we develop a long-term memory consolidation mechanism that selectively retains hard and discriminative samples, ensuring robust knowledge retention. To facilitate research in this area, we construct and release two multi-modal and textual benchmarks, \texttt{MMLongCL-Bench} and \texttt{TextLongCL-Bench}, providing a valuable resource for evaluating long-term CL approaches.
Experimental results show that \texttt{Long-CL} outperforms the previous state-of-the-art by 7.4\% and 6.5\% AP on the two benchmarks, respectively, demonstrating the effectiveness of our approach.
\end{abstract}

\vspace{-1mm}
\section{Introduction}
\label{sec:intro}
\vspace{-1mm}
Continual learning (CL) has emerged as a crucial paradigm in machine learning, aiming to equip models with the ability to learn sequentially from a stream of tasks over time~\cite{parisi2019continual}. This setup mimics human learning, where knowledge is acquired incrementally while preserving previously learned information. However, a significant challenge in CL is catastrophic forgetting—a phenomenon where a model's performance on earlier tasks deteriorates as it learns new ones~\cite{french1999catastrophic,kirkpatrick2017overcoming}. While traditional CL settings address this issue within a limited number of tasks, real-world scenarios often demand systems capable of handling an extensive sequence of tasks. This necessitates a shift toward~\textbf{long-term} CL, where models must scale to accommodate a vast number of tasks over prolonged periods.

In long-term CL, the challenges of catastrophic forgetting are exacerbated due to the sheer volume of tasks and the extended duration of learning.  
To reduce the catastrophic forgetting problem, existing approaches can be broadly categorized into several groups: regularization-based, architecture-based, rehearsal-based, and prompt-based methods~\cite{yang2024recent}.
For the regularization-based methods, EWC~\cite{kirkpatrick2017overcoming} introduces a method that computes the importance of parameters for past tasks using Fisher information, O-LoRA~\cite{wang2023orthogonal} constrains model updates for different tasks within mutually orthogonal low-rank subspaces.
For the architecture-based methods, CL-MoE~\cite{huai2025cl} introduces a dual-momentum expert framework, which selects experts effectively and performs momentum updates.
For the rehearsal-based methods, EMR~\cite{wang2019sentence} employs a working memory mechanism to strategically rehearse stored samples.
For the prompt-based methods, L2P~\cite{wang2022learning} introduces a prompt-based CL, where task-specific prompts are learned to adapt a pre-trained model to new tasks without modifying its backbone.  

\begin{wrapfigure}{r}{0.5\linewidth}  
  \centering
  \vspace{-4mm}  
  \includegraphics[width=\linewidth]{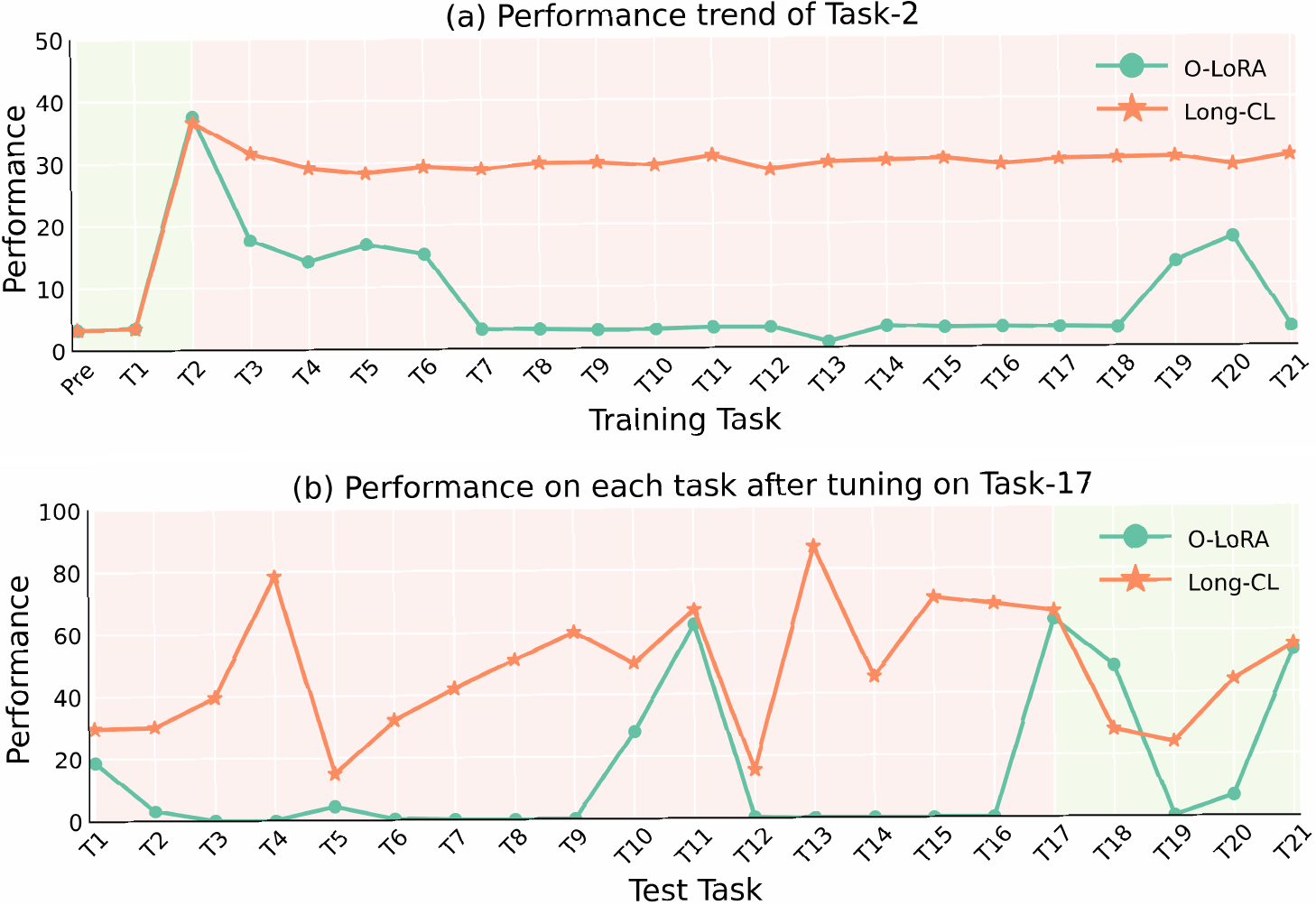}
  \vspace{-7mm}
  \caption{The performance of our method and O-LoRA on \texttt{MMLongCL-Bench}.}
  \label{intro}
  \vspace{-3mm}
\end{wrapfigure}

Although previous studies have proven effective, most existing methods perform CL under a few task types with relatively short data streams. In contrast, real-world scenarios always require long-term CL, which requires models to continually learning across heterogeneous task types and non-stationary, long-term data streams. 
There are two primary challenges for long-term CL:
\textbf{First}, as the number of tasks grows, catastrophic forgetting becomes more severe. Our preliminary experiments reveal that existing CL approaches struggle to maintain performance in such extended settings. Taking Figure~\ref{intro}(a) as an example, the knowledge of task 2 is forgotten after task 7, failing to retain knowledge effectively over time.
It is important to address the questions: ``What knowledge should be learned, and how should memory be updated?" 
\textbf{Second}, there is a notable absence of comprehensive benchmarks tailored for long-term CL, limiting the ability to evaluate and compare methods under realistic conditions. 

To address these challenges, we formalize the long-term CL task and propose a human memory mechanism-inspired framework, \texttt{Long-CL}, for long-term continual learning, which efficiently manages and consolidates information over time. Our approach introduces a task-core memory management strategy that efficiently indexes and updates crucial memories as learning progresses. Additionally, we develop a long-term memory consolidation mechanism that selectively retains hard and discriminative samples, ensuring robust knowledge retention without excessive memory overhead. 
We also construct and release two comprehensive benchmarks: \texttt{MMLongCL-Bench} (multi-modal) and \texttt{TextLongCL-Bench} (textual) to foster research in this emerging area. These benchmarks serve as valuable resources for evaluating long-term CL approaches under realistic conditions.
Through extensive experiments on both benchmarks, we demonstrate that \texttt{Long-CL} significantly outperforms strong baselines, showcasing its effectiveness in mitigating catastrophic forgetting and achieving superior performance in long-term CL scenarios. 
Our main contributions can be concluded as follows:
\begin{itemize}[leftmargin=*, align=left]
    \item We introduce a long-term continual learning task, where the model learns from a sequence of vast datasets over time. We also propose two long-term continual learning benchmarks: \texttt{MMLongCL-Bench} for vision-language tasks and \texttt{TextLongCL-Bench} for text-only tasks.
    \item We propose \texttt{Long-CL}, a memory-inspired approach with task-core memory management (MemMan) and long-term memory consolidation (MemCon) mechanisms. MemMan indexes the task-specific crucial memory and updates it dynamically via the relationships between the current and previous tasks. MemCon identifies hard and differential samples to reinforce long-term memory.
    \item We conduct extensive experiments on \texttt{MMLongCL-Bench} and \texttt{TextLongCL-Bench} and analyze the performance of general CL baselines. Also, our \texttt{Long-CL} achieves state-of-the-art performance by comparing with the strong baselines, demonstrating the effectiveness of our method. 
\end{itemize}
\vspace{-1mm}
\section{Related Work}
\label{sec:related}
\vspace{-1mm}
\textbf{Large Language Models.} 
Large Language Models (LLMs)~\cite{devlin2019bert,brown2020language,raffel2020exploring,sun2021ernie,le2023bloom,achiam2023gpt} have revolutionized the field of NLP by achieving remarkable performance across a variety of tasks. More recent developments include PaLM~\cite{chowdhery2023palm}, which demonstrates strong multilingual and reasoning capabilities, and LLaMA~\cite{touvron2023llama}, which provides high-performance models in an open-source format. Additionally, Gemini~\cite{team2023gemini} introduces a multimodal approach, handling various input types beyond text. These advancements underscore the rapid evolution and diversification of LLMs in recent years.
Recent research on Vision Language Models (VLMs) has flourished~\cite{grattafiori2024llama,li2024ocean,deitke2024molmo,anil2023gemini,wu2024next,chen2024internvl}, with the emergence of GPT-4o~\cite{hurst2024gpt} refreshing our understanding of VLMs. LLaVA~\cite{liu2023vision} initially updates only the projection matrix for feature alignment pretraining, and subsequently fine-tunes the projection matrix and the language model end-to-end. Qwen2.5-VL~\cite{bai2025qwen2} employs dynamic resolution processing and absolute temporal encoding, enabling flexible adaptation to images of varying sizes and long-duration videos.
In this paper, we investigate the application of LLMs/VLMs within a CL setting, aiming to enable the acquisition of novel knowledge while preserving previously learned information.

\textbf{Continual Learning for LLMs.} 
In the rapidly iterating landscape of knowledge, LLMs must emulate human-like learning capabilities, assimilating new knowledge while preserving previously acquired information. 
However, LLMs often suffer from catastrophic forgetting~\cite{kirkpatrick2017overcoming,yang2024recent} when confronted with continuous and non-stationary data streams, leading to significant declines in performance on earlier tasks and a reduction in overall generalization ability.
Numerous studies~\cite{lopez2017gradient,sun2019lamol,song2023conpet,wang2023orthogonal,chen2023lifelong,liu2023class,jha2024clap4clip,kim2024vlm} have attempted to address the forgetting problem in LLMs.
ER~\cite{rolnick2019experience} maintains the model's performance on previous tasks by replaying a subset of data from previously learned tasks during the training of new tasks.
LWF~\cite{li2017learning} employs knowledge distillation to preserve performance on previously learned tasks without requiring access to old task data.
DIKI~\cite{tang2024mind} designs a residual attention mechanism to inject new task knowledge into the frozen VLM model through the residual path, avoiding modifying the original attention path.
CL-MoE~\cite{huai2025cl} introduces RMoE to construct hybrid expert routing weights to achieve fine-grained expert allocation for different tasks and samples, and MMoE distinguishes experts into task-shared and task-specific types, and updates them in a dynamic momentum way. 
Most existing methods perform continual instruction tuning on a monotonous task type and data distribution, with short-lived data streams.
Unlike previous studies, we propose two long-term CL benchmarks that incorporate multiple task types, diverse data distributions, and a long-horizon data stream, which is more aligned with real-world CL scenarios.

\begin{table}[t!]
\vspace{-3mm}
\centering
\caption{Comparison with existing CL datasets. ABSC, TC, TPC, and TS mean aspect-based sentiment classification, text classification, text pair classification, and text summarization.}
\label{table: comparison of datasets}
\setlength{\tabcolsep}{1.4mm}{
\begin{tabular}{llccccc}
\hlineB{4}
& Benchmark          & \#Dataset  & \#Task  & \#Task Type  &  \#Train & \#Test \\ \hline
\multirow{5}{*}{Multimodal} & IMNER/IMRE \cite{chen2024continual}  & 10  & 1 & MSA   & 22.0k & 1.2k  \\
&CLOVE~\cite{lei2023symbolic}                 & 12   & 1      & VQA   & 235.5k  &  35.4k \\ 
&VQACL \cite{zhang2023vqacl}           & 10  & 1 &  VQA   & 537.5k & 23.5k  \\
&CLiMB~\cite{srinivasan2022climb}      & 4          & 2     & VQA,VE   & 136.6k  &  24.7k \\ 
& \texttt{MMLongCL-Bench}   & 21      & 4   &  VQA, MSA, VE, TR   & 503.0k  & 92.6k \\ \midrule
\multirow{5}{*}{Textual} & ABSC \cite{ke-etal-2021-adapting}  & 19        & 1    &  ABSC & 10.1k & 23.3k \\
& LNT-Benchmark~\cite{wang2023orthogonal} & 15 & 3    & TC, TPC, QA  & 55.7k   & 70.9k \\
& ClassificationCL \cite{de2019episodic} &     5       &    1  &  TC  & 575.0k  & 38.0k \\ 
& LFPT5 \cite{qinlfpt5} & 9  &  3  & TC, NER, TS &  610.6k   &  68.9k  \\
&\texttt{TextLongCL-Bench} & 30   & 3 &  NER, RE, EE & 397.3k & 39.3k \\
\hlineB{4}
\end{tabular}}
\vspace{-3mm}
\end{table}

\vspace{-1mm}
\section{Preliminaries}
\label{sec:Preliminaries}
\vspace{-1mm}
\subsection{Long-term Continual Learning Benchmark}

\textbf{Vision-Language Long-term Continual Learning Benchmark.}
The proposed benchmark \texttt{MMLongCL-Bench} for long-term CL contains 21 datasets characterized by different task types and data distributions. These datasets are CLEVR~\cite{johnson2017clevr}, DVQA~\cite{kafle2018dvqa}, GQA~\cite{hudson2019gqa}, ImageNet~\cite{deng2009imagenet}, InfographicVQA~\cite{mathew2022infographicvqa}, IST~\cite{karatzas2015icdar}, SemEval~\cite{sharma2020semeval}, NLVR~\cite{suhr2017corpus}, NLVR2~\cite{suhr2017corpus}, P2GB~\cite{chen2024plug}, ScienceQA~\cite{lu2022learn}, SMART-101~\cite{cherian2023deep}, SNLI-VE~\cite{xie2019visual}, t4sa~\cite{vadicamo2017cross}, twitter2015~\cite{zhang2018adaptive}, twitter2017~\cite{yu2020improving}, VCOPA~\cite{yeo2018visual}, VCR~\cite{zellers2019recognition}, VizWiz~\cite{gurari2018vizwiz}, VQA v2~\cite{goyal2017making}, VSR~\cite{liu2023visual}. For task variety, \texttt{MMLongCL-Bench} spans a wide range of tasks, including Visual Question Answering (VQA), Visual Reasoning (VR), Sentiment Analysis (SA), and Text Recognition (TR). Regarding data diversity, it spans a broad spectrum of visual sources such as document layouts, charts, tables, posters, blog content, puzzles, etc. 

\textbf{Textual Long-term Continual Learning Benchmark.}
We also collect a textual benchmark, \texttt{TextLongCL-Bench}, which includes 30 datasets that consists of Named Entity Recognition (NER), Relation Extraction (RE) and Event Extraction (EE)~\cite{wang2023instructuie}. These datasets are ACE, ADE, AnatEM, bc2gm, bc4chemd, bc5cdr, BroadTweet, conll03, conll04, CrossNER-AI, CrossNER-literature, CrossNER-music, CrossNER-politics, CrossNER-science, FabNER, FindVehicle, GENIA, GIDS, HeavyNER, kbp37, mit-movie, mit-restaurant, ncbi, New-York-Times, NYT11, Ontonotes, SciERC, TweetNER7, WikiANN, WikiNeural. Please refer to the supplementary material for more details.

\subsection{Comparison with Existing CL Datasets and Methods}

\textbf{Comparison with Existing CL Datasets.}
We compare our benchmarks with conventional CL benchmarks (Table~\ref{table: comparison of datasets}). 
\texttt{MMLongCL-Bench} covers 21 datasets and 503k training samples across four multimodal tasks, which outperforms existing multimodal CL datasets in terms of dataset size and task diversity. \texttt{TextLongCL-Bench} encompasses 30 datasets for complex textual tasks that are crucial for understanding complex semantic relationships within text, making it valuable for NLP research.
This supports comprehensive evaluations and fair comparisons of long-term CL techniques. 

\begin{figure}[t]
\centering
\includegraphics[width=1\columnwidth]{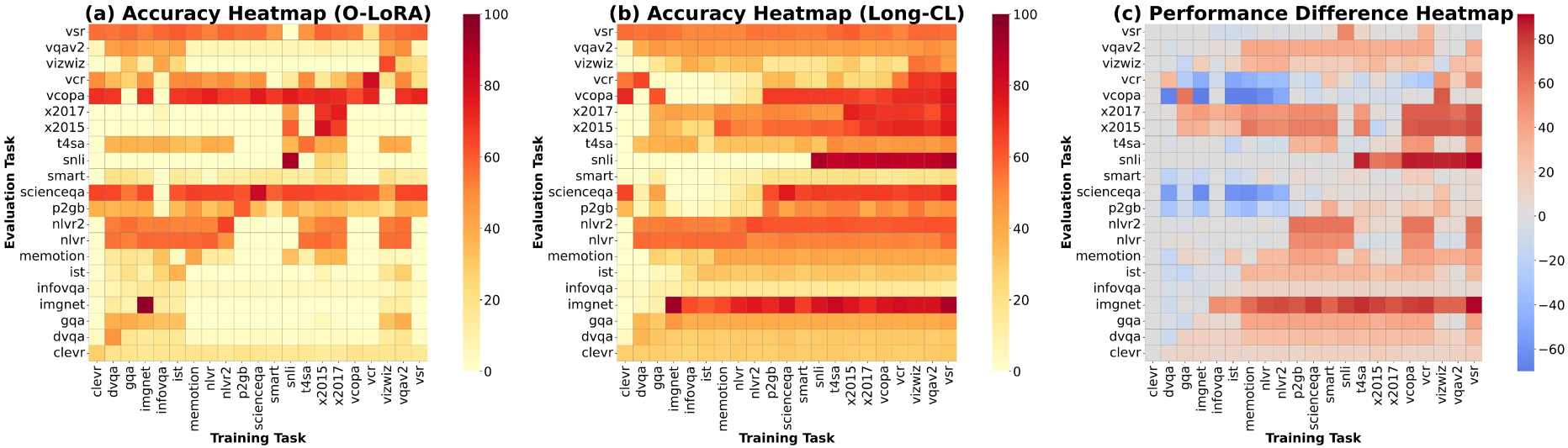}
\vspace{-6mm}
\caption{The performance heatmap of our method and O-LoRA on \texttt{MMLongCL-Bench}. The horizontal axis indicates the training order and the vertical axis represents the performance after tuning.}
\label{pre}
\end{figure}

\textbf{Comparison with Existing CL Methods.}
First, existing methods struggle to effectively retain knowledge from previous tasks when scaling up to long sequences of tasks. 
As shown in Figure~\ref{intro}(b), after fine-tuning on task 17, the model almost completely forgets the knowledge acquired from previous tasks.
Additionally, as shown in Figure~\ref{pre}(b), when the model is trained on the final task, the performance on previous tasks almost drops to zero. In the region below the sub-diagonal, it can be observed that the model performs poorly in most cases, with accuracy generally below 20\%.
Second, we calculate the performance difference of \texttt{Long-CL} and O-LoRA in Figure~\ref{pre}(c). The region below the sub-diagonal predominantly shows red, indicating \texttt{Long-CL} superior resistance to forgetting compared to the baseline. The area above the sub-diagonal reflects zero-shot capability, where \texttt{Long-CL} remains largely comparable to the baseline, with only minor occasional declines. This demonstrates that our model can better balance knowledge transfer between new and old tasks, thereby effectively reducing the forgetting effect for long-term CL.

\subsection{Task Definition}
In this work, we formalize \textbf{long-term continual learning} for LLMs as a paradigm requiring sustained adaptation over an \textit{extremely large sequence of tasks}, mimicking real-world deployment where knowledge evolves dynamically over months or years. 
It is unlike traditional CL, which is typically constrained to shorter task sequences with limited heterogeneity.  
Specifically, the benchmark contains $M \gg 1$ datasets with different data distributions and task types, denoted by task identifiers $\{T_1, T_2,..., T_M\}$.
The $T_t$ task includes its specific training data $D_{t} = \{(X^{t}_{i}, Y^{t}_{i})\}^{N_{t}}_{i=1}$ with $N_{t}$ data tuples, where $X$, $Y$ denotes the instruction and answer respectively.  
The goal is to continuously incorporate new knowledge while maintaining performance on previously seen tasks over a long sequence of learning stages.
In the evaluation phase, the model is required to make predictions for unseen test instances without access to task identity, highlighting the importance of task-agnostic generalization and memory consolidation in long-term CL.

\begin{figure*}[!t]
\centering
\includegraphics[width=1\textwidth]{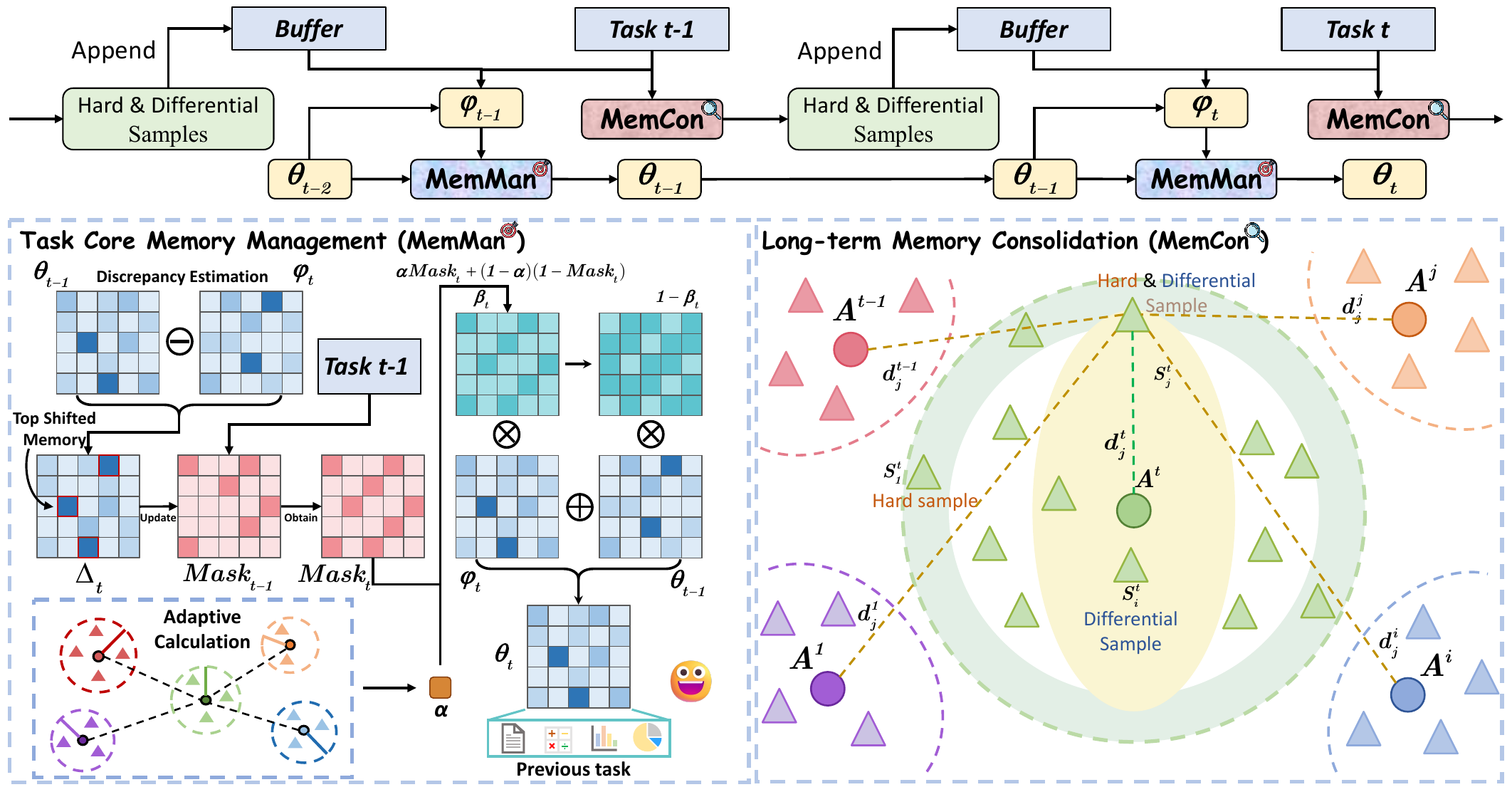}
\vspace{-4mm}
\caption{The framework of our \texttt{Long-CL}. First, we introduce MemMan to index the task-core memory and adjust the model's memory adaptively based on the relationship between the current and previous tasks. Then, MemCon learns the key knowledge using hard sample selection and differential sample selection to choose task-relevant informative (i.e., the samples in the green annulus) and cross-task generalizable samples (i.e., the samples in the yellow annulus), respectively.
}
\label{framework}
\end{figure*}

\section{Method}
\label{sec:method}
In this section, we propose a human memory mechanisms-inspired framework \texttt{Long-CL} for long-term continual learning, as shown in Figure~\ref{framework}. \texttt{Long-CL} consists of two effective components, Task-Core Memory Management (MemMan) and Long-term Memory Consolidation (MemCon).
First, we introduce MemMan to position the task-specific memory and update it dynamically by calculating the relationships between the current task and previous tasks.
Subsequently, we introduce MemCon, which aims to reinforce long-term memory via the hard and differential samples.

\subsection{Overview}
Large language models (LLMs) obtain language understanding, generation, and reasoning capabilities with rich knowledge by pre-training on large-scale corpora.
To make full use of LLMs, we unify various tasks under a generative paradigm, enabling LLMs to produce appropriate outputs conditioned on given instructions.
We regard the corresponding instruction $X$ as input, and the LLMs compute the probability of the complete answer $Y$ in an autoregressive manner.
Specifically, given a vision-language task instruction template ``$<image>$ Is any of the pizza missing? Answer the question using a single word or phrase." with the ground truth answer $Y$ (i.e., ``yes" or ``no"), we optimize the model using the negative log-likelihood loss:
\begin{equation}
  \mathcal{L} = -\sum^{L}_{j=1} \log p_{\theta}(Y_{j}|X, Y_{1:j-1}),
  \label{eq1}
\end{equation}
where $L$ denotes the total number of tokens in the answer $Y$, $Y_{1:j-1}$ represents the subsequence of tokens preceding position $j$, and $Y_j$ denotes the token at the $j_{th}$ position. $\theta$ corresponds to the set of trainable parameters in the LLMs.
Notably, in consideration of parameter efficiency and computational cost, we adopt the Low-Rank Adaptation (LoRA) technique \cite{DBLP:conf/iclr/HuSWALWWC22} during fine-tuning. Specifically, LoRA adapters are injected into the feed-forward neural network (FFN) layers within each Transformer block, while the original model parameters are kept frozen.

Let $\theta_{t-1}$ be the parameters of LoRA block trained on task $\{1, 2, ..., {t-1}\}$ and $\varphi_t$ as the parameters learned by $t$ task base on $\theta_{t-1}$.
At task $t$, we model the data distribution of $D_t$ and derive the associated prototype representation, represented as $A^t$. 
MemMan finds the task-core memory by computing the differences between memory units in $\varphi_t$ and $\theta_{t-1}$, identifies the top-K units that contribute most to task $t$, and logs their respective indices. Furthermore, MemMan adaptively derives a fusion weight $\alpha$ by evaluating the similarity of the prototype between task $t$ and previous tasks, enabling the integration of current and historical memory into a unified $\theta_t$ that captures both current and past knowledge.
Subsequently, MemCon introduces a memory consolidation mechanism for experience replay optimization, thereby improving model retention and generalization in long-term continual learning scenarios.
First, it employs a hard sample selection mechanism to identify and prioritize informative outliers—typically challenging samples—for task $t$. Second, to preserve knowledge from previous tasks, MemCon introduces a differential sample selection strategy that retrieves instances most representative of historical task distributions.

\subsection{Task-Core Memory Management}
Human learning leverages a fundamental mechanism where individuals continuously acquire new information and integrate it with existing knowledge structures. This dynamic process of memory management and conceptual synthesis represents a key cognitive advantage, enabling humans to generalize patterns, make logical inferences, and adapt to novel situations.
We argue that LLMs should resemble humans not only in learning behavior but also in knowledge management. After acquiring new knowledge, it is crucial to ensure compatibility with existing knowledge and to reshape prior memory accordingly. To address this issue, we introduce the Task-Core Memory Management (MemMan), which consists of Task-Core Memory Indexing and Adaptive Memory Updating, aiming to identify critical memory locations and perform adaptive memory reconstruction.

\textbf{Task-Core Memory Indexing.}
This module aims to identify task-core memory by computing a discrepancy estimation between the previous model $\theta_{t-1}$ and the current model $\varphi_t$, where $\varphi_t$ is obtained through instruction tuning of $\theta_{t-1}$ on task $t$. Specifically, we estimate each memory unit shift degree between $\theta_{t-1}$ and $\varphi_t$, and identify the top-K units with the most significant deviations, as formulated below:
\begin{equation}
  \mathcal{I}_t = \text{TopK}\left( \left\{ d_i = \left\| \theta_{t-1}^{(i)} - \varphi_t^{(i)} \right\|_2 \right\}_{i=1}^N, \, K \right)
  \label{eq2}
\end{equation}
where $d_i$ is the Euclidean shift distance of the $i_{th}$ unit, $\text{TopK}(\cdot, K)$ denotes the selection of the top-K indices with the largest drift values. $\mathcal{I}_t$ represents the set of memory unit indices selected in task $t$.

We regard $\mathcal{I}_t$ as the key memory units that contribute the most to task $t$. These indices are stored in a mask matrix $\text{Mask}_t$, which is dynamically updated during continual fine-tuning across different tasks.
\begin{equation}
  \text{Mask}_t^{(i)} = 
\begin{cases}
1, & \text{if } \text{Mask}_{t-1}^{(i)} = 1 \text{ or } i \in \mathcal{I}_t \\
0, & \text{otherwise}
\end{cases}
  \label{eq3}
\end{equation}
where $\text{Mask}_t^{(i)}$ denotes the $i_{th}$ position in $\text{Mask}_t$. $\text{Mask}_t$ will be used in the subsequent adaptive memory updating process.

\textbf{Adaptive Memory Updating.}
In this part, we update the memory adaptively by calculating the magnitude according to the similarity between the current task and the previous tasks.
To obtain the prototype representation $A^{t}$ of task $t$, we first extract features for each sample $X$ using an encoder $f$. We then compute the task-specific prototype by performing mean pooling over all joint representations within the task $t$:
\begin{equation}
A^t = \frac{1}{n_t} \sum_{i=1}^{n_t} f(X)
  \label{eq4}
\end{equation}
where $n^t$ denotes total samples of task $t$. Meanwhile, we can obtain $\{A^{i}\}^{t-1}_{i=1}$ from previous task. To dynamically adjust the influence of current versus past memory, we introduce an adaptive weighting factor $\alpha_t$. This weight is calculated based on the relative distance between the current prototype and historical prototypes, normalized by the total inter-prototype distances:
\begin{equation}
\alpha_t = \frac{\sum_{j=1}^{t-1} \left\| A^t - A^j \right\|_2}{\sum_{1 \leq i < j \leq t-1} \left\| A^i - A^j \right\|_2}
  \label{eq5}
\end{equation}
Intuitively, the adaptive weight $\alpha_t$ reflects the semantic novelty of the current task relative to prior tasks. When the task is semantically distant from prior ones (\eg, in the early stages of continual learning), $\alpha_t$ tends to be large, encouraging stronger updates on memory units. As more tasks are introduced, $\alpha_t$ gradually decays, promoting more stable updates. To avoid overly conservative memory updates that could hinder the integration of new knowledge, we impose a lower bound $ \alpha_{\min} = \lambda$. This ensures that a certain level of plasticity is maintained.

To perform fine-grained and position-aware memory fusion that considers both task-level memory importance and memory-wise relevance across tasks, we construct a weight matrix $ \beta \in \mathbb{R}^d$. This matrix incorporates the adaptive weight $\alpha_t$, along with the memory importance mask $ \text{Mask}_t $, formally:
\begin{equation}
\beta_t = \alpha_t \cdot \text{Mask}_t + (1 - \alpha_t) \cdot (\boldsymbol{1} - \text{Mask}_t)
  \label{eq6}
\end{equation}
This design ensures that memories identified as important receive higher attention, especially when $\alpha_t$ is large. As $\alpha_t$ decreases over time, updates to key positions will become increasingly conservative, enhancing stability for long-term knowledge retention.
We perform adaptive memory update by combining the current model \( \varphi_t \) and the previous model \( \theta_{t-1} \) in an element-wise weighted manner:
\begin{equation}
\theta_t = \beta_t \cdot \varphi_t + (1 - \beta_t) \cdot \theta_{t-1}
  \label{eq6}
\end{equation}
where $\theta_t$ denotes the updated model for task $t$, balancing new knowledge injection and accumulated memory preservation through position-aware weighting.

\subsection{Long-term Memory Consolidation}
Our MemMan module integrates task-relevant knowledge at the parameter level through adaptive memory fusion to update the long-term memory. However, not all samples contribute equally to long-term knowledge retention. High-value and high-difficulty samples, such as rare patterns, decision boundary cases, or previously misclassified examples, are crucial in mitigating forgetting. To address this issue, we introduce a Long-term Memory Consolidation module (MemCon), which contains the hard and differential sample selection strategies, aiming to prioritize task-relevant informative samples and cross-task generalizable samples.

\textbf{Hard Sample Selection.} 
To enhance the model’s capability in handling current task challenges, we design a hard sample selection strategy based on semantic distance to the task prototype. Specifically, we aim to retain samples that are most dissimilar to the current task’s prototype representation $A^t$, as these samples often correspond to hard cases, outliers, or underrepresented decision boundaries.
For each training sample \( s^t_i \in D_t \), we compute its Euclidean distance to the task prototype \( A^t \):
\begin{equation}
d^t_i = \left\| s^t_i - A^t \right\|_2, \quad
\mathcal{H}_t = \text{TopR}_{h} \left( \left\{ (s^t_i, d^t_i) \mid s^t_i \in D_t \right\} \right)
  \label{eq7}
\end{equation}
We select the $R_h\%$ samples with the largest $d_i$ values into the $\mathcal{H}_t$, which represents the selected hard sample set for task $t$.

\textbf{Differential Sample Selection.} 
In addition to hard samples that expose the model’s uncertainty on the current task, it is also essential to retain samples that are semantically consistent across tasks. These samples help reinforce shared representations and reduce distributional drift in continual learning.
To this end, we introduce a Differential Sample Selection strategy, which identifies samples that are globally aligned with prior task prototypes. Specifically, for each sample \( s^t_i \in D_t \), we compute the cumulative Euclidean distance to all previous task prototypes:
\begin{equation}
z^t_i = \sum_{j=1}^{t-1} \left\| s^t_i - A^j \right\|_2, \quad 
\mathcal{G}_t = \text{MinR}_{g} \left( \left\{ (s^t_i, z^t_i) \mid s^t_i \in D_t, \min_{j=1,\dots,t-1} \left\| s^t_i - A^j \right\|_2 \geq \delta \right\} \right)
  \label{eq8}
\end{equation}
To avoid selecting samples that are overly biased toward any single task, we further impose a local distance constraint. Specifically, we define a minimum distance threshold $\delta$ and discard samples that are too close to any individual task prototype $A^j$, ensuring that selected samples reflect global consistency rather than local dominance. We select the minimum $R_g\%$ samples with the smallest cumulative distance values as the differential sample set $\mathcal{G}_t$.

The final replay buffer for task $t$ is constructed by combining both sets, $\mathcal{R}_t = \mathcal{H}_t \cup \mathcal{G}_t$.
This dual-strategy selection allows the model to retain both task-difficult and task-general samples, improving both task adaptability and long-term consistency.
\section{Experimental Results}
\label{sec:exp}
\subsection{Experimental Setups}
\paragraph{Dataset and Evaluation Metrics.}

We conduct experiments on our \texttt{MMLongCL-Bench} and \texttt{TextLongCL-Bench} benchmarks. 
We adopt two widely used metrics for continual learning evaluation~\cite{lopez2017gradient,chaudhry2018riemannian}: Final Average Performance ($AP$) and Average Forgetting ($AF$). 
$AP$ quantifies the model’s ability to retain previously acquired knowledge after completing the continual fine-tuning process. Formally, given $m_{j,k}$ as the test performance on task $k$ after training up to task $j$, the metric is defined as $AP = \frac{1}{M} \sum^{M}_{i=1} m_{M,i}$, where $M$ denotes the total number of tasks. $AF$ measures the degradation in performance on earlier tasks caused by learning new ones. It is calculated as the average drop in accuracy from the time a task was learned to the end of the learning sequence: $AF = \frac{1}{M-1} \sum^{M-1}_{i=1} m_{i,i} - m_{M,i}$. We use accuracy (\%) as the evaluation measure for $m$.

\begin{table}[!t]
\centering
\small
\caption{Performance (\%) of our \texttt{Long-CL} and distinct continual learning methods on \texttt{MMLongCL-Bench} and \texttt{TextLongCL-Bench}. The best results are emphasized in \textbf{bold}.}
\label{tab2}
\setlength{\tabcolsep}{1.2mm}{
\begin{tabular}{lcccccc|ccccc}
\hlineB{4}
\multirow{2}{*}{Methods}                            & \multicolumn{4}{c}{MMLongCL-Bench}                                & \multirow{2}{*}{$AP(\uparrow)$} & \multicolumn{1}{l|}{\multirow{2}{*}{$AF(\downarrow)$}} & \multicolumn{3}{l}{TextLongCL-Bench}                                      & \multirow{2}{*}{$AP(\uparrow)$} & \multicolumn{1}{l}{\multirow{2}{*}{$AF(\downarrow)$}} \\ \cline{2-5} \cline{8-10}
                                                           & VQA            & VE             & MSA            & TR             &                                 & \multicolumn{1}{l|}{}                                  & \multicolumn{1}{c}{RE} & \multicolumn{1}{c}{EE} & \multicolumn{1}{c}{NER} &                                 & \multicolumn{1}{l}{}                                  \\ \hline
Vanilla                              & 12.34          & 27.92          & 0.00              & 0.30           & 15.35         & 28.42              & 34.33                       & 32.07                        &  47.89                      & 44.20                           &  15.55   \\
O-LoRA~\cite{wang2023orthogonal}     & 14.33          & 27.97          & 0.00              & 0.20           & 16.12         & 28.36              & 34.16                       & 29.02                        &  45.36                      & 42.20                           &  17.58   \\
LWF~\cite{li2017learning}            & 27.04          & 34.41          & 42.83             & 24.30          & 32.72         & 10.28              & 30.90                       & 35.03                        &  48.81                      & 44.18                           &  15.57    \\
MoELoRA~\cite{chen2024coin}          & 25.76          & 36.58          & 21.19             & 0.00           & 27.79         & 15.40              & 35.33                       & 37.18                        &  47.55                      & 44.35                           &  15.40    \\
EWC~\cite{kirkpatrick2017overcoming} & 37.40          & 42.85          & 44.00             & \textbf{31.00} & 40.43         & 2.33               & 33.69                       & 29.82                        &  46.49                      & 42.95                           &  16.88    \\
ER~\cite{rolnick2019experience}      & 35.88          & 48.94          & 54.61             & 8.30           & 43.11         & -0.74              & 43.50                       & 53.33                        &  56.80                      & 53.58                           &  5.89     \\
I-LoRA~\cite{li2025analyzing}        & 35.16          & 49.17          & 52.73             & 20.70          & 43.16         & -0.73              & 43.47                       & 54.72                        &  56.44                      & 53.36                           &  6.07     \\
CL-MoE~\cite{huai2025cl}             & 32.39          & 55.67          & 55.94             & 6.90           & 44.53         & -2.21              & 38.07                       & 36.04                        &  47.67                      & 45.04                           &  14.64    \\ \hline
\texttt{Long-CL}                     & \textbf{45.74} & \textbf{57.53} & \textbf{60.36}    & 23.00          & \textbf{51.93}& \textbf{-9.93}     &  \textbf{49.71}             & \textbf{60.12}               &  \textbf{63.43}             & \textbf{60.12}                  &  \textbf{-0.89} \\
Multitask                            & 48.91          & 63.04          & 57.19             & 29.70          & 54.96         & -                  &  55.46                      &  65.42                       &  63.21                      & 61.48                           &  -        \\ \hlineB{4}
\end{tabular}}
\end{table}

\textbf{Baselines.}
To demonstrate the effectiveness of our method, we select several typical continual learning methods. The replay-based method ER~\cite{rolnick2019experience}, I-LoRA~\cite{li2025analyzing}, regularization-based methods O-LoRA~\cite{wang2023orthogonal}, EWC~\cite{kirkpatrick2017overcoming}, and LWF~\cite{li2017learning}, architecture-based methods MoELoRA~\cite{chen2024coin}, CL-MoE~\cite{huai2025cl}.
Multitask represents the performance of the model that trains on all the tasks once, while Vanilla indicates the performance of the model trained on a sequence of tasks without using any methods.

\textbf{Implementation Details.}
We use LLaVA-7B~\cite{liu2024visual} for vision-language tasks and Qwen2.5-7B~\cite{yang2024qwen2} for textual tasks. 
For all baseline methods, we follow the implementation details and configurations from the original papers to ensure faithful reproduction. For the rehearsal method, we set the memory buffer size to 20\% of the training dataset. For architecture-based methods, we set the expert number to 4. During training, we train each task for one epoch with a batch size of 16. For hyperparameter settings, we set $K$ = 10\%, $\alpha_{min}$ = 0.3, $R_g$ = $R_h$ = 10\% and $\delta$ = 0.8*$D_{max}$ / 2, where $D_{max}$ represent the maximum distance of prototypes. We use AdamW as the optimizer with the learning rate of 1$e^{-4}$, and employ the cosine learning rate scheduler.


\subsection{Main Results}
We report the performance of the baselines and our \texttt{Long-CL} method on \texttt{MMLongCL-Bench} and \texttt{TextLongCL-Bench}, as shown in Table~\ref{tab2}. Due to page width limitations, we group the 21 tasks in \texttt{MMLongCL-Bench} into four broader categories and the 30 tasks in \texttt{TextLongCL-Bench} into three categories, using the average accuracy within each category for evaluation. Please refer to the supplementary for complete results. 
Based on the experimental results, we draw the following conclusions.
\textbf{First}, our \texttt{Long-CL} achieves the state-of-the-art on both \texttt{MMLongCL-Bench} and \texttt{TextLongCL-Bench}, demonstrating its effectiveness and task-agnostic nature.
\textbf{Second}, compared to Vanilla LLaVA and Qwen2.5, our \texttt{Long-CL} improved $AP$ by approximately 36.58\% (51.93\% vs. 15.35\%) and 15.92\% (60.12\% vs. 44.20\%) on \texttt{MMLongCL-Bench} and \texttt{TextLongCL-Bench}, consistently yields notable gains on every task. For $AF$, \texttt{Long-CL} improves the performance by approximately 38.35\% (-9.93\% vs. 28.42\%) and 16.44\% (-0.89\% vs. 15.55\%). Notably, our $AF$ value is -9.93\% and -0.89\%, indicating that the average performance on the previous tasks surpasses their respective fine-tuning results, which demonstrates strong backward transfer capability.  
\textbf{Third}, compared to language tasks, vision-language tasks suffer more severe catastrophic forgetting under the long-term continual learning setting. We attribute this to the fact that vision-language tasks exhibit more discrete data distributions and larger inter-task discrepancies.
\textbf{Fourth}, compared with the upper bound method Multitask that trains on the merged datasets of the benchmark, our \texttt{Long-CL} approaches the performance of multitask learning, though a gap remains. We aim to explore more effective approaches for long-term continual learning in future work.


\begin{wraptable}{r}{0.6\textwidth}
\centering
\small
\vspace{-4mm}
\setlength{\tabcolsep}{0.5mm}{
\begin{tabular}{l|cc|cccc|cc}
\hlineB{4}
  & \multicolumn{2}{c|}{Method} & \multicolumn{4}{c|}{MMLongCL-Bench}                       & \multirow{2}{*}{AP} & \multirow{2}{*}{AF} \\ \cline{1-7}
  & MemMan       & MemCon       & VQA            & VE             & MSA            & TR             &                     &                     \\ \hline
a & \ding{55}     & \ding{55}     & 12.34           & 27.92          & 0.00              & 0.30           & 15.35               & 28.42               \\
b & \ding{51}      & \ding{55}     & 31.01          & 47.74          & 48.85          & 12.70          & 39.91               & 2.31                \\
c & \ding{55}     & \ding{51}      & 39.94          & 50.09          & 58.57          & \textbf{24.80}          & 46.63               & -4.90                 \\
d & \ding{51}      & \ding{51}      & \textbf{45.74} & \textbf{57.53} & \textbf{60.36} & 23.00 & \textbf{51.93}      & \textbf{-9.93}      \\ 
\hlineB{4}
\end{tabular}}
\vspace{-1mm}
\caption{The results of ablation study on \texttt{MMLongCL-Bench}.}
\label{tab3}
\vspace{-2mm}
\end{wraptable}

\subsection{Ablation Study}
To evaluate the effectiveness of each component in \texttt{Long-CL}, we conduct an ablation study for \texttt{Long—CL} on \texttt{MMLongCL-Bench}, as shown in Table~\ref{tab3}.
Specifically, we conduct experiments with MemMan only, MemCon only, and the complete configuration incorporating both components. 
By comparing variants (a,b) and (a,c), we infer that both MemMan and MemCon contribute significantly to long-term continual learning in vision-language tasks. Specifically, MemMan enables more adaptive memory updates by leveraging parameter changes and the guidance of the $\text{Mask}$ during task transitions. MemCon enhances memory consolidation by selecting task-relevant informative samples and cross-task generalizable samples, and reinforcing them through a rehearsal mechanism. Moreover, we observe that MemCon outperforms the MemMan strategy, as partial forgetting may still occur even with effective memory management in MemMan. In contrast, MemCon alleviates this issue to some extent through its replay mechanism. 
By analyzing (a, d), we observe that the combination of MemMan and MemCon effectively alleviates forgetting in long-term continual learning and enhances knowledge transfer across tasks. 
The ablation study on textual tasks is similar to that on vision-language tasks, please refer to the supplementary material for the result.


\subsection{Further Analysis}
\begin{figure*}[t!]
\vspace{-3mm}
    \centering
    \begin{minipage}[b]{0.4\textwidth}
        \centering
        \includegraphics[width=\linewidth]{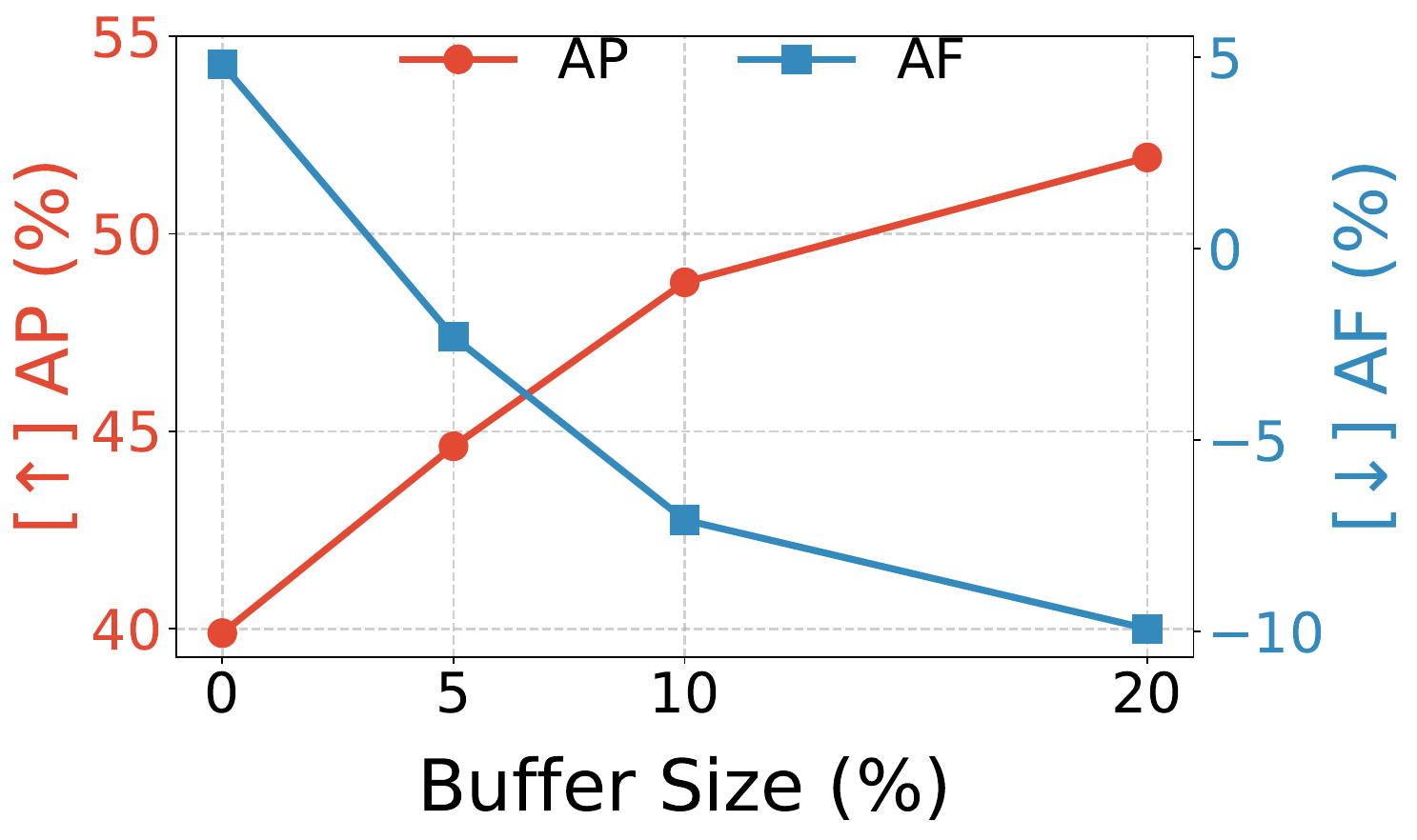}
    \end{minipage}
    \begin{minipage}[b]{0.4\textwidth}
        \centering
        \includegraphics[width=\linewidth]{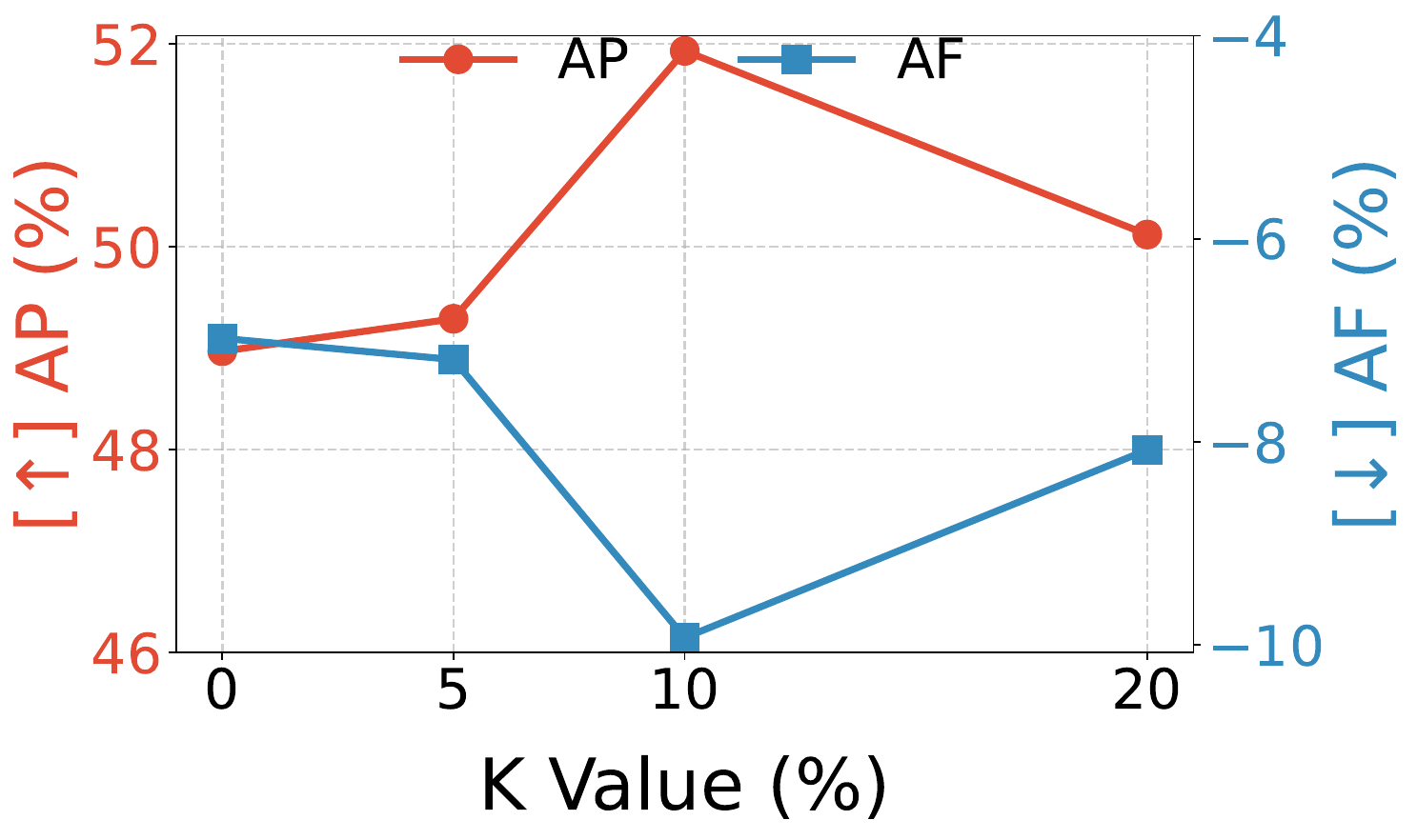}
    \end{minipage}
    \vspace{-3mm}
    \caption{Performance(\%) of \texttt{Long-CL} with different buffer size and $K$ value on \texttt{MMLongCL-Bench}.}
    \label{fig:hyperparameters}
    \vspace{-3mm}
\end{figure*}

\textbf{Impact of Buffer Size $R_t$.} To investigate the impact of buffer size $R_t$, we conduct additional experiments on \texttt{MMLongCL-Bench}, as shown in Figure~\ref{fig:hyperparameters}. The results show that our \texttt{Long-CL} achieves significant performance gains even with only a 5\% replay buffer. As the buffer size $R_t$ increases, performance improves persistently, though training cost also increases. When $R_t$ is set to 20\%, our \texttt{Long-CL} reaches 94.5\% of the multitask performance, striking a favorable balance between mitigating forgetting and resource efficiency. For further analysis of text tasks conducted on \texttt{TextLongCL-Bench}, please refer to the supplementary material.


\textbf{Impact of Hyperparameter $K$.} 
We investigate the impact of hyperparameters $K$ in MemMan, as shown in Figure~\ref{fig:hyperparameters}. The results indicate that, compared to using Adaptive Memory Updating alone (\ie, $K$ = 0), incorporating Task-Core Memory Indexing helps identify critical memory locations and leads to more stable updates. Moreover, choosing an appropriate $K$ value is crucial. A small $K$ may lead to the loss of task-critical memory, while a large $K$ will reduce model plasticity, making it harder to adapt to subsequent tasks. Therefore, in our experimental setup, we set $K$ to 10\%. 

\begin{wraptable}{r}{0.45\textwidth}
\small
\vspace{-4mm}
\setlength{\tabcolsep}{0.8mm}{
\begin{tabular}{l|cl|cc|cc}
\hlineB{4}
\multirow{2}{*}{Method} & \multicolumn{2}{c|}{Order1}     & \multicolumn{2}{c|}{Order2} & \multicolumn{2}{c}{Order3}                            \\
                        & AP    & \multicolumn{1}{c|}{AF} & AP           & AF           & AP                        & AF                        \\ \hline
CL-MoE                  & 44.53 & -2.21                   & 40.25        & 1.82        & 39.15                     & 3.44                     \\
Long-CL                 & 51.93 & -9.93                   & 49.16        & -7.15        & \multicolumn{1}{l}{48.60} & \multicolumn{1}{l}{-6.45} \\ \hlineB{4}
\end{tabular}}
\vspace{-1mm}
\caption{The results of different task orders.}
\label{tab6}
\vspace{-3mm}
\end{wraptable}

\textbf{Impact of Task Order.} 
We investigate the impact of different task orders on \texttt{Long-CL}. Specifically, we randomly shuffled the task order in \texttt{MMLongCL-Bench} and conducted long-term CL, the results are shown in Table~\ref{tab6}. Experimental results show that our method consistently outperforms the strong baseline CL-MoE under three different task orders, demonstrating its effectiveness. Furthermore, whereas models typically suffer from varying levels of forgetting under different task orders, \texttt{Long-CL} exhibits more stable performance, highlighting its robustness to task sequence variations.

\begin{wraptable}{r}{0.45\textwidth}
\small
\setlength{\tabcolsep}{0.8mm}{
\begin{tabular}{c|cccc|cc}
\hlineB{4}
\multirow{2}{*}{$\alpha_t$} & \multicolumn{4}{c|}{MMLongCL-Bench}                               & \multirow{2}{*}{AP} & \multirow{2}{*}{AF} \\ \cline{2-5}
                       & VQA            & VE             & MSA            & TR             &                     &                     \\ \hline
0.3                    & 44.75          & 53.83          & 59.70          & \textbf{25.20} & 50.12               & -8.08               \\
0.5                    & 44.56          & 56.08          & 59.89          & 21.00          & 50.75               & -8.73               \\
0.7                    & 44.88          & 56.66          & 59.05          & 22.20          & 50.99               & -9.02               \\
adaptive               & \textbf{45.74} & \textbf{57.53} & \textbf{60.36} & 23.00          & \textbf{51.93}               & \textbf{-9.93}      \\ \hlineB{4}
\end{tabular}}
\vspace{-1mm}
\caption{The results of different $\alpha_t$ value.}
\label{tab7}
\vspace{-3mm}
\end{wraptable}

\textbf{Impact of Hyperparameter $\alpha_t$.} 
We investigate the impact of the hyperparameter $\alpha_t$ on \texttt{Long-CL}, with results shown in Table~\ref{tab7}. Experimental results suggest that using a fixed hyperparameter $\alpha_t$ may not be ideal. If $\alpha_t$ is too large, it may lead to excessive loss of critical memory in later stages. On the contrary, the model may fail to retain useful knowledge from each task. Adopting Adaptive Memory Updating allows the model to perform more aggressive memory updates in the early stages and preserve key information in later stages, which is beneficial for long-term continual learning.

\vspace{-1mm}
\section{Conclusions and Further Works}
\label{sec:conclusion}
\vspace{-1mm}
In this paper, we introduce a more realistic long-term continual learning setting and propose two benchmarks: \texttt{MMLongCL-Bench} for vision-language tasks and \texttt{TextLongCL-Bench} for text-only tasks. To address the forgetting problem in long-term continual learning, we propose the \texttt{Long-CL} framework, which incorporates two key components: MemMan and MemCon. Inspired by human learning mechanisms, MemMan identifies critical memory locations and performs adaptive memory updates. MemCon strengthens knowledge retention by replaying a combination of task-relevant informative and cross-task generalizable samples. Extensive experiments on \texttt{MMLongCL-Bench} and \texttt{TextLongCL-Bench} demonstrate that our \texttt{Long-CL} approach achieves state-of-the-art performance. Ablation and additional studies further validate the effectiveness of our method. In the future, we aim to integrate tasks from different modalities, making the setting more realistic and further addressing the forgetting problem faced by LLMs in long-term continual learning.

{
    \small
    \bibliography{main}
}

\end{document}